\pdfoutput=1
\documentclass[11pt]{article}

\usepackage[final]{acl}

\usepackage{times}
\usepackage{latexsym}
\usepackage{colortbl} % 在导言部分导入colortbl包

\usepackage[T1]{fontenc}
\usepackage[utf8]{inputenc}

\usepackage{microtype}

\usepackage{inconsolata}

\usepackage{graphicx}
\usepackage{multirow}
\usepackage{amsmath}
\usepackage{amssymb}
\usepackage{mathtools}
\usepackage{amsthm}
\usepackage{microtype}
\usepackage{subcaption}
\usepackage{booktabs} % for professional tables
\usepackage{enumitem} % 导言区加载

\usepackage{adjustbox} % 用于图片垂直居中对齐
\usepackage{array}     % 用于表格增强

\usepackage{fancyhdr}
\usepackage{authblk}
\usepackage{pifont}

\fancypagestyle{plain}{ % 定义简单页码样式
    \fancyhf{} % 清除所有默认的页眉页脚
    \fancyhead[L]{ % 左侧页眉
        \raisebox{-0.3\height}{\includegraphics[height=20pt]{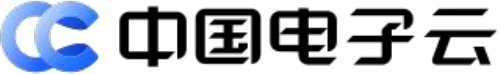}} % Logo
    }
}

\title{Global Context Compression with Interleaved Vision-Text Transformation}

\author{
  \textbf{Dian Jiao}\textsuperscript{*}\textsuperscript{\dag},
  \textbf{Jiaxin Duan}\textsuperscript{*},
  \textbf{Shuai Zhao},
  \textbf{Jiabing Leng},
  \textbf{Yiran Zhang},
  \textbf{Feng Huang}\textsuperscript{\ddag} \\  %  
  China Electronics Cloud Technology Co., Ltd. \\
  \texttt{\{jiaodian, duanjiaxin, zhaoshuai, lengjiabing, zhangyiran, huangfeng01\}@cestc.cn}
}

\begin{document}
\maketitle

\renewcommand{\thefootnote}{\fnsymbol{footnote}} % 将脚注编号改为符号形式

\footnotetext[1]{Equal contributions.} % 对应上标 *（符号序号1）
\footnotetext[2]{Project leader.}
\footnotetext[3]{Corresponding author.}
% 注意：\footnotetext[1] 中的数字对应符号编号，1表示*

% \footnotetext{* Equal contributions.}
% \footnotetext{\dag Project leader.}

\begin{abstract}
Recent achievements of vision-language models in end-to-end OCR point to a new avenue for low-loss compression of textual information.
This motivates earlier works that render the Transformer's input into images for prefilling, which effectively reduces the number of tokens through visual encoding, thereby alleviating the quadratically increased Attention computations.
However, this partial compression fails to save computational or memory costs at token-by-token inference.
In this paper, we investigate global context compression, which saves tokens at both prefilling and inference stages.
Consequently, we propose VIST2, a novel Transformer that interleaves input text chunks alongside their visual encoding, while depending exclusively on visual tokens in the pre-context to predict the next text token distribution.
Around this idea, we render text chunks into sketch images and train VIST2 in multiple stages, starting from curriculum-scheduled pretraining for optical language modeling, followed by modal-interleaved instruction tuning.
We conduct extensive experiments using VIST2 families scaled from 0.6B to 8B to explore the training recipe and hyperparameters. 
With a 4$\times$ compression ratio, the resulting models demonstrate significant superiority over baselines on long writing tasks, achieving, on average, a 3$\times$ speedup in first-token generation, 77\% reduction in memory usage, and 74\% reduction in FLOPS.
Our codes and datasets will be public to support further studies.
\end{abstract}

\section{Introduction}
\begin{figure}[!t]
\centering
\includegraphics[width=0.95\linewidth]{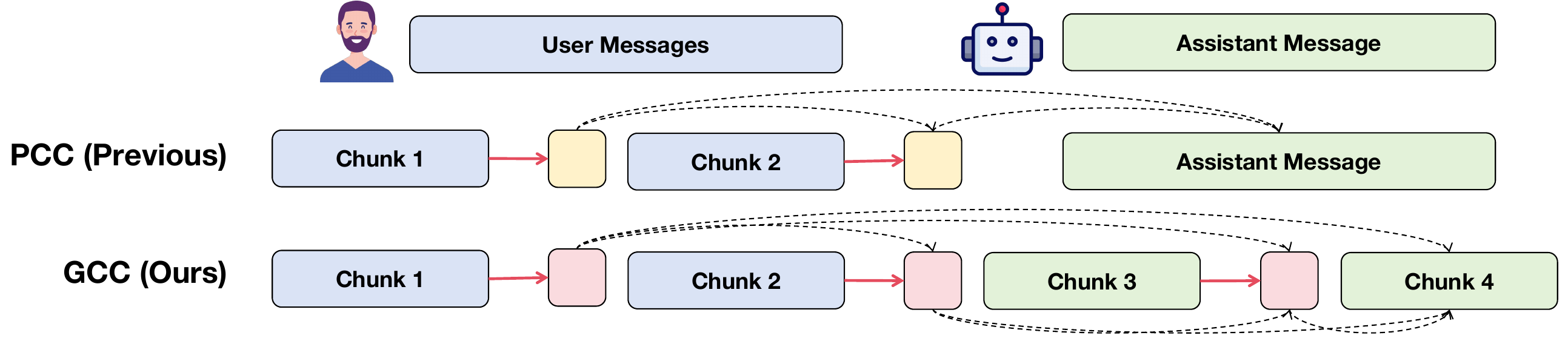}
\caption{
The illustration of context compression of the Transformer.
\textcolor{red}{Right arrow} indicates transforming a text chunk to its latent representation. 
}
\label{fig:1}
\end{figure}

% Large language models (LLMs) with Transformer architecture suffer from a significant context limitation due to the self-attention that has a $\mathcal{O}(n^2)$ computational complexity. This yields an intensive demand for \textit{context compression} to reduce computing costs.
Large language models (LLMs) with Transformer architecture face significant challenges in context scaling because the complexity of self-attention increases quadratically ($\mathcal{O}(n^2)$) with sequence length. This yields an urgent need for \textit{context compression} that can reduce computing costs without sacrificing model performance.
% Context compression approaches are proposed to tackle this challenge.
Existing approaches for context compression are \textit{sparse attention} and \textit{hierarchical encoding}.
Following informatics theory, sparse attention~\cite{li2025admtree,Longformer,Sparser} drops out detected tokens with marginal information to reduce the Attention operators. In contrast, hierarchical encoding~\cite{Glyph,c3} (in Figure~\ref{fig:1}) splits a long text into ordered chunks, where each is compressed into densely informative representations.
By preserving the complete context, hierarchical encoding effectively prevents information loss, thereby earning widespread interest.
% There are mainly two approaches to address this problem.
% The first one is sparse Attention mechanisms. 
% For example, LongFormer and X-Former mask redundant tokens with low information density using zero attention scores, which effectively reduces the computing cost.
% % , whereas the input of LLMs remains long.
% Another way is hierarchical Transformer. As illustrated in Figure 1, hierarchical Transformer splits a long text into ordered chunks, and each chunk is compressed using an additional LLM or visual encoder, or block-wise causal attention to several latent tokens, thereby significantly truncating the model input.
% and employ which extends the standard Transformer with an external encoder that compresses long input tokens into several latent embeddings. 
% However, these approaches suffer from a low compression rate, and the encoder structure is not unified.
Recently, the emerging vision-language models (VLMs), such as dots.ocr~\cite{dots.ocr}, MinerU-VLM~\cite{MinerU}, and DeepSeek-OCR~\cite{DeepSeek} demonstrate remarkable performance in optical character recognition (OCR). 
The key behind their success - \textit{optical compression}~\cite{text-or-pixels} opened a new door for a more promising hierarchical encoding.
For example, Glyph~\cite{Glyph} renders millions of tokens into images and achieved a 4$\times$ lossless compression based on a powerful visual encoder. 

Despite these achievements, the mentioned approaches equip LLMs with partial rather than global context compression. 
Taking the human-AI conversation in Figure 1 as an example, human queries are compressed in prefilling, while the AI responses are generated token-by-token without any compression. 
In the case of long-text generation, the model performs intensive computations during training and requires significant memory to store KV-Caches during inference, resulting in considerable costs.
To address this problem, we conduct a pioneering exploration for global context compression, aiming to save tokens at both prefilling and inference stages.
Specifically, we propose VIST2, a novel Transformer that interleaves text chunks and their optical encoding in input, and predicts the next text token conditioned on visual tokens in pre-context.
% Specifically, we propose VIST2, a novel Transformer that interleaves visual-text alternation in the forward pass, and predicts the next text token conditioned on visual tokens already in context.
This interleaved visual-text transformation, named \textit{Optical Language Modeling} (OLM), effectively bridges the gap between partial and global compressions but is often overlooked in existing works.
% VIST2 adopts a sandwich structure with modern VLMs, compressing a visual encoder and an LLM connected by a linear layer.
In experiments, VIST2 is implemented using Qwen3 and SigLip2 connected by a linear projection layer. 
We pretrain this model through several stages, starting with image captioning and OCR to warm up the visual modules, followed by the OLM to adjust the backbone LLM.
We then fine-tune it on conversations with long queries and responses, regularized by a modal-interleaved chat template.
Extensive results show that our VIST2, with a compression ratio of 4:1, achieves a 3$\times$ speedup in the first-token delay while also reducing memory usage and FLOPS during inference by over 75\% each.
% Extensive results demonstrate that with a 4$\times$ compression ratio, our model achieves a 3$\times$ speedup in the first-token delay, saving 25\% in memory usage and 25\% in FLOPS during inference.

\section{Background and Problem Definition}
This work focuses on text-to-text generation - the general task form of modern linguistic intelligence, which facilitates human-machine interaction through conversations.
Given a long input text $\mathbf{X} = [x_1, x_2, ..., x_L]$ with the target $\mathbf{Y} = [y_1, y_2, ..., y_M]$, the standard Decoder-only LLMs, e.g., GPT, Llama, Qwen, etc, aim to model the conditional probabilities via next-token prediction: 
\begin{equation}
\label{eq:1}
P(\mathbf{Y}|\mathbf{X}) \to \sum_{i}^{M} \mathcal{P}_\theta(y_i|\mathbf{X},y_{<i})
\end{equation}
In the settings where $L$ and $M$ exceed the effective context window of standard transformers, the computational complexity increases quadratically, resulting in intensive costs that hinder the training of large language models (LLMs).

\textbf{Partial Context Compression} (PCC) methods address this problem by partitioning a long input $\mathbf{X}$ into $n$ continuous chunks $\{\mathcal{C}_i\}_{i=1}^n$, where each chunk $\mathcal{C}_i = [x^i_1, ..., x^i_k]$ contains $k$ tokens. 
Additionally, a \textbf{text renderer} $\mathcal{R}(\cdot)$ is employed to render each chunk $\mathcal{C}_i$ into a grayscale optical image $\mathcal{V}_i \in \mathbb{R}^{H \times W \times 3}$. Consequently, they convert the causal language modeling objective in Eq.~\ref{eq:1} into visual-language modeling:
\begin{equation}
\label{eq:2}
P(\mathbf{Y}|\mathbf{X}) \to \sum_{i}^{M} \mathcal{P}_\theta(y_i|\{\mathbf{v}_k\}_{k=1}^{L},y_{<i})
\end{equation}
where $\mathbf{v}_k=VisualEncoder(\mathcal{V}_i)$ are visual tokens derived from the chunk $\mathcal{C}_i$.

% \textbf{Global Context Compression} (GCC). 
Although PCC extends the context window of Transformers by up to 10 times wider, the visual compression of text tokens works only for prefilling without the support of inference. As a result, PCC enables long-text understanding (LTU), while still facing challenges in long-text generation (LTG), such as storytelling, novel writing, and complicated multi-step reasoning. 
To address this limitation, we propose \textit{global context compression}, which enables both LTU and LTG through interleaved text-vision compression and vision-text modeling. 
% % Specifically, we concatenate $\mathbf{X}$ and $\mathbf{Y}$ into a single text $\mathbf{U}=$, 
% For illustration convenience, we let $\mathbf{U}=\{\hat{\mathcal{C}}_i\}_{i=1}^m$ be an ultra context that consists of the concatenated input and target texts. Our goal is formally:
% \begin{equation}
% \label{eq:2}
% P(\mathbf{Y}|\mathbf{X}) \to \sum_{i} \mathcal{P}_\theta(u^{i}_{j}|\hat{\mathbf{v}}_{<i},u^{i}_{<j})
% \end{equation}
% where $\hat{\mathbf{v}}_i$ are visual tokens of the chunk $\hat{\mathcal{C}}_i=\{u^{i}_{j}\}_{j=1}^{K}$, and $K=|\hat{\mathcal{C}}_i|$ is the chunk size.
% We propose a novel Transformer architecture to achieve GCC, which is detailed in the next section.

\section{Method: VIST2}
\textbf{VIST2} is an efficient large language model (LLM) architecture that achieves global context compression by performing iterative visual-text transformations. The core idea behind VIST2 is to utilize the spatial redundancy of rendered text to compress information into a dense visual latent space.

% our goal is to enable efficient processing while preserving critical long-range dependencies. We address this by compressing textual chunks into compact visual representations. 
% % The challenge is to derive a compressed representation $\mathbf{v}_i \in \mathbb{R}^{m \times d}$ for each chunk such that $m \ll k$ while retaining sufficient semantic information for contextual reasoning.
% A . 

\begin{figure}
\centering
\includegraphics[width=0.95\linewidth]{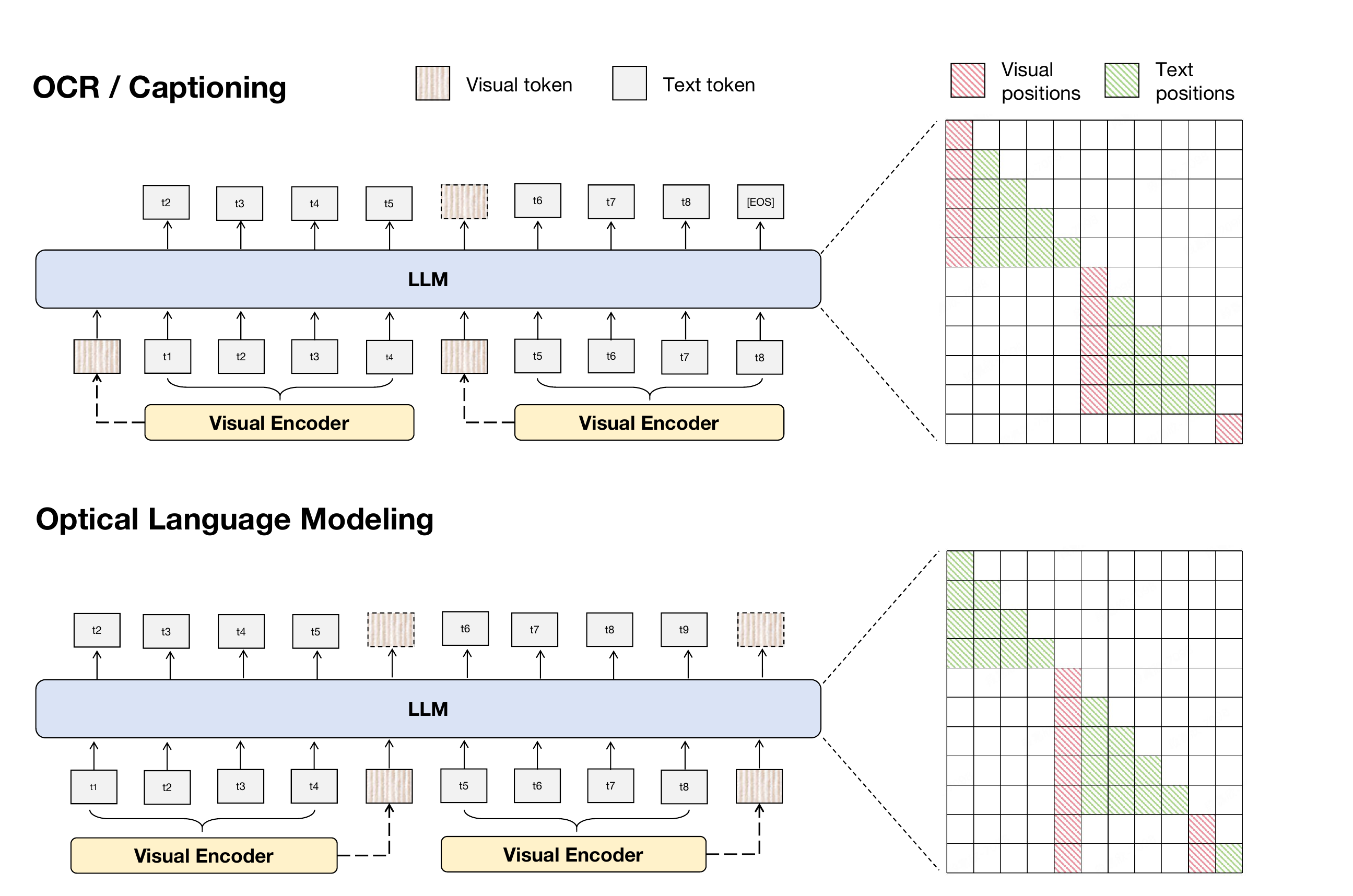}
\caption{The illustration of pre-training.}
\label{fig:2}
\end{figure}

\subsection{Model Architecture}

As illustrated in Figure 1, VIST2 features a sandwich architecture, a popular design in VLMs, comprising a visual encoder (VE) and an LLM backbone connected by a modal aligner. 
% Notably, we discrete the dense attention of LLM to save computation complexity.

% We employ a monospace font with fixed character grid to ensure consistent token-to-pixel mapping, preserving spatial relationships within the chunk. Rendering parameters (font size, line spacing) are selected to maximize information density while remaining legible to the visual encoder.

\noindent \textbf{Visual Encoder} $\mathcal{E}(\cdot)$ processes images rendered from structure-free texts into token embeddings. For this reason, we utilize a pretrained Vision Transformer with the patches sized 16$\times$16 (ViT-L/16), which yields $m = \lfloor \frac{H}{16} \rfloor \times \lfloor \frac{W}{16} \rfloor$ visual tokens per image:
% $$
% \tilde{\mathbf{v}}_i = \mathcal{E}(\mathcal{R}(\mathcal{\hat{C}}_i)) \in \mathbb{R}^{m \times d_v}
% $$
$$
\tilde{\mathbf{v}}_i = \mathcal{E}(\mathcal{\hat{V}}_i) \in \mathbb{R}^{m \times d_v}
$$

\noindent \textbf{Modal Aligner} $\mathcal{M}(\cdot)$ aligns the outputs of VE with the LLM's embedding space, which is achieved through a multilayer perceptron:
$$
\hat{\mathbf{v}}_i = \tanh(\tilde{\mathbf{v}}_i W_m + b_m) \in \mathbb{R}^{m \times d_{lm}}
$$
where $W_m \in \mathbb{R}^{d_v \times d_{lm}}$ and $b_m \in \mathbb{R}^{d_{lm}}$ are trainable parameters.

\noindent \textbf{Sparse Attention Mechanism}. The final input to the LLM is an interleaved sequence:
$$
\mathbf{H} = [\hat{\mathbf{c}}_1, \hat{\mathbf{v}}_1, \hat{\mathbf{c}}_2, \hat{\mathbf{v}}_2, ..., \hat{\mathbf{v}}_{n-1}, \hat{\mathbf{c}}_n]
$$
where $\hat{\mathbf{c}}_i$ are token embeddings of $\hat{\mathcal{C}}_i$. 
% To achieve global context compression, we perform a sparse casual attention to constrain tokens visibility and save computational cost.
To achieve GCC, we implement a sparse causal attention that constrains token visibility and saves computations.
The \textbf{attention mask} is illustrated in Figure 2 (lower), where each token's visibility follows:
$$
\text{Mask}(q, k) = 
\begin{cases}
1 & \text{if } k \in \{\mathbf{c}_i, \tilde{\mathbf{v}}_{<i}\} \text{ for } q \in \tilde{\mathbf{v}}_i \\
\text{Causal} & \text{otherwise}
\end{cases}
$$
This ensures that visual tokens act as contextual memory accessible to subsequent chunks, but the context itself can not pass to future textual content. Additionally, the \textbf{position} of an embedded token converts to:
$$
% \text{Pos}(h_j) = 
% \begin{cases}
% \sum|\hat{\mathbf{v} }_{<i}|+j & \text{if } h_j \in \mathbf{c}_i \\
% \sum|\hat{\mathbf{v} }_{<i}|+|\mathbf{c}_{i-1}|+j & \text{if } h_j \in \mathbf{v}_i
% \end{cases}
\text{Pos}(j) = \sum|\hat{\mathbf{v} }_{<i}|+j
$$
where $|\hat{\mathbf{v}}_i|$ denotes the number of visual tokens in the $i$-th preceding chunk, and $j$ represents the local offset within the current chunk. 
In this way, both encoding of visual and text tokens share a continuous and modality-agnostic positional space, which is crucial for saving computations in iterative vision-text transformation.
% In this way, visual tokens retain their positional information while being processed in relation to their respective chunks, thereby achieving a structured representation that allows for efficient retrieval and integration of visual data with textual content without accessing future information.

\subsection{Model Training}
The training of VIST2 encounters two main challenges: first, the LLM and VE are well-trained, while the connector is initialized from scratch, resulting in asynchronous optimization of model parameters. Additionally, the modifications to attention layers pose further challenges over the standard LLM fine-tuning.
To address these challenges, we propose a multi-stage training recipe, including staged pre-training and instruction-based fine-tuning. Table 1 reports the details.

\subsubsection{Pre-training}
% We first pretrain VIST2 on the \textbf{OLM} objective using synthetically generated document-image pairs. 
We first pretrain VIST2 for image captioning to warm up the modal aligner, with VE and LLM frozen. 
Subsequently, we train VIST2 with a multi-turn OCR (MT-OCR) task to enable VE for text compression. Specifically, we flip the adjacent odd and even positions in the LLM input: $\mathbf{H} \to \tilde{\mathbf{H}}=[\hat{\mathbf{v}}_1, \hat{\mathbf{c}}_1, \hat{\mathbf{v}}_2, \hat{\mathbf{c}}_2, ..., \hat{\mathbf{v}}_{n}, \hat{\mathbf{c}}_n]$. 
MT-OCR asks the model to recover the content of text chunks conditioned on their optimal features, by minimizing the following training loss:
% For each text chunk $\mathcal{T}_i$ sampled from our corpus, we render its corresponding image $\mathcal{V}_i$ and construct interleaved sequences $\mathbf{X}_{\text{olm}} = [\mathcal{V}_1, \mathcal{T}_1, ..., \mathcal{V}_n, \mathcal{T}_n]$.
$$
% \frac{1}{\sum_{i=1}^n |\hat{\mathcal{C}}_i|} 
\mathcal{L}_{ocr} = -\sum_{i=1}^n \sum_{j=1}^{|\hat{\mathcal{C}}_i|} \log \mathcal{P}_\theta(u^i_j \mid \{\hat{\mathcal{V}}_{k}\}_{k<i}, \hat{\mathcal{C}}_{<i}, u^i_{<j})
$$

During this stage, we update the parameters of both VE and the modal aligner with only the LLM frozen.
To enhance training convergence, we implement a curriculum schedule for the second stage. It consists of three difficulty levels: easy, which involves OCR of a single image; medium, which encompasses OCR of 2 to 4 images; and hard, which requires OCR of more than 4 images. 
% This task forces the model to: (1) extract textual content from rendered images, (2) establish cross-modal alignment, and (3) learn to utilize visual tokens as reliable memory cues. We pretrain for 100K steps with batch size 512 using a learning rate of 1e-4.

After the above two stages, VIST2 is capable of: 1) compressing format-free texts into images with a high compression rate, and 2) recovering the essential texts' information from their optical images. 
A further step is required to fit the LLM with sparse attention tailored for the OLM objective in Eq.2. The loss function is:
% We call this optical language modeling (OLM) to differentiate from the standard single-modal LM. This is achieved by optimizing the loss function with frozen visual modules:
% During fine-tuning, we process genuine long-context inputs $\mathbf{X} = [\mathcal{C}_1, \mathcal{C}_2, ..., \mathcal{C}_n]$ by: d
% 1. Rendering each chunk to obtain $\{\mathcal{V}_i\}_{i=1}^{n-1}$
% 2. Constructing the interleaved sequence $\mathbf{X}_{\text{ft}} = [\mathcal{C}_1, \mathcal{V}_1, ..., \mathcal{C}_n]$
% 3. Optimizing the standard language modeling loss:
% \frac { 1 } { \sum _ { i=1 } ^ { n } | \hat { \mathcal { C } } _i| } 
\[
\mathcal { L } _ { olm } = - \sum _ { i } ^ { n } \sum _ { j=1 } ^ { | \hat { \mathcal { C } } _i| } \log \mathcal { P } _ \theta (u^i_j \mid \{ \hat { \mathcal { V } } _ { k } \} _ { k<i } , u^i_ { <j } )
\]
% Crucially, we freeze the visual encoder after pre-training and only update the mapping layer and LLM parameters during fine-tuning to preserve the learned visual-textual alignment while adapting to downstream tasks. We employ LoRA adapters on attention modules for parameter-efficient fine-tuning.
Figure 2 visualizes the attention masks used in pretraining, conditioned on stages.

\subsubsection{Supervised Fine-tuning}
We fine-tune VIST2 with \textit{modal-interleaved instruction tuning} to align with real-world applications.
% We consider both long-writing tasks, where instructions are commonly concise while responses are relatively long, and long-input tasks, where queries are cumbersome while the responses are relatively short (e.g., the answers to needle-in-a-haystack benchmarks are just a single letter indicating the true option). 
This process covers two primary scenarios: (1) long-writing tasks, characterized by concise instructions and extensive narrative responses, and (2) long-context tasks, which involve cumbersome queries but brief outputs (e.g., single-letter answers in "needle-in-a-haystack" benchmarks). 
Given instruction data in single-turn conversations, we compress the query and response independently, i.e., chunks of size $K$ are encoded into $\beta$ visual tokens. To handle the sequence tail, a residual segment of length $m$ is compressed if $m > \beta$; otherwise, it remains in its raw tokenized form to preserve fine-grained information. 
Mathematically, we train the model to minimize the following loss function:
\[
\mathcal{L} = - \sum _ { i } ^ { n } \sum _ { j=1 } ^ { | \hat { \mathcal { C } } _i| } \log \mathcal { P } _ \theta (u^i_j \mid \{ \hat { \mathcal { V } } _ { k } \} _ { k<i } , u^i_ { <j }, V(x))
\]
In this equation, $V(x)$ is an input query compressed following the mentioned principle, and the cross-entropy loss is solely sourced from the OLM of the response.
Our ablation study reveals that leveraging the pre-trained parameters significantly accelerates convergence on these challenging instruction-following tasks. 
% Detailed hyperparameters and dataset statistics are provided in Appendix~\ref{sec:appendix}.
Please see Appendix~\ref{sec:appendix} for a more comprehensive understanding of the orchestration of our training pipeline.  
% \subsection{Submitting Papers}

% \subsubsection{Instruction Tuning}
% To align VIST2 with real-world applications, we perform modal-interleaved instruction tuning. We curate a diverse dataset covering two primary scenarios: (1) long-writing tasks, characterized by concise instructions and extensive narrative responses, and (2) long-context tasks, which involve cumbersome queries but brief outputs (e.g., single-letter answers in "needle-in-a-haystack" benchmarks). 
% For single-turn conversation data, we compress the query and response independently: segments of size $K$ are encoded into $\beta$ visual tokens. To handle the sequence tail, a residual segment of length $m$ is compressed if $m > \beta$; otherwise, it remains in its raw tokenized form to preserve fine-grained information. Leveraging the pre-trained weights from the OLM (Original Language Model) significantly accelerates convergence on these challenging alignment tasks. Detailed hyperparameters and dataset statistics are provided in Appendix~\ref{sec:appendix}.

\section{Experiment}
\label{sec:3}

% \begin{table}[!ht]
% \centering
% \scriptsize
% \caption{Architecture of VIST2-family models.}
% \begin{tabular}{ccccc}
% \toprule
% \multirow{2.5}{*}{\textbf{Model}} & \multicolumn{2}{c}{\textbf{Visual Encoder}} & \textbf{Aligner} & \textbf{LLM} \\
% \cmidrule{2-3}\cmidrule{4-4}\cmidrule{5-5}
% & \textbf{Patch} & \textbf{Resolution} & \textbf{Hidden} & \textbf{Parameters} \\
% \midrule
% VIST2-0.6B & 16$\times$16 & Naive & 19,456 & 0.6B \\
% VIST2-4B & 16$\times$16 & Naive & 19,456 & 4B \\
% VIST2-8B & 16$\times$16 & Naive & 19,456 & 8B \\
% \bottomrule
% \end{tabular}
% \label{tab:m}
% \end{table}

\subsection{Experimental Settings}
We implement VIST2 using open-source models: with SigLIP2~\cite{SigLIP} as the visual encoder and Qwen3~\cite{qwen3} as the backbone LLM. The connector is a randomly initialized multilayer perceptron, with a hidden dimension size of 19,456. 
In experiments, we keep the visual modules while scaling the size of Qwen3 from 0.6B to 8B, resulting in a family of VIST2 models, named VIST2-0.6B, VIST2-4B, and VIST2-8B. 
% Table~\ref{tab:m} reported detailed architectural hyperparameters.
Our experiments are conducted on 8$\times$Nvidia H200 GPUs. 
Refer to Appendix 1 for the detailed configuration of datasets and hyperparameters.
% Table~\ref{tab:data} summarizes the datasets alongside training stages, and Table~\ref{tab:hp} reports hyperparameters.  

\subsection{Pre-training Performance}
\label{sec:3-1}
\textbf{Image Captioning}.
We pretrain VIST2 on the image captioning task using 690 million open-source samples collected from SA1B\footnote{\url{https://www.modelscope.cn/datasets/Tongyi-DataEngine/SA1B-Dense-Caption}} and CoCo-CN~\cite{coco-cn}.  
Since the captioning task converges easily, we terminate the training after the loss stabilizes at $\sim$0.9. Then, we compare the resulting models with powerful VLMs. Table~\ref{tab:2} reports the evaluation results on the CoCo test set, with ROUGE scores and CLIPScore (CS) as the matrices.
VIST2 performed equivalently with the competitors, indicating the efficient training of the connector.

\begin{figure*}[!t]
\centering
% \begin{subfigure}[b]{0.32\linewidth}
% \includegraphics[width=\linewidth]{images/loss_plot.png}
% \caption{Captioning training loss.}
% \end{subfigure}
\begin{subfigure}[b]{0.32\linewidth}
\includegraphics[width=\linewidth]{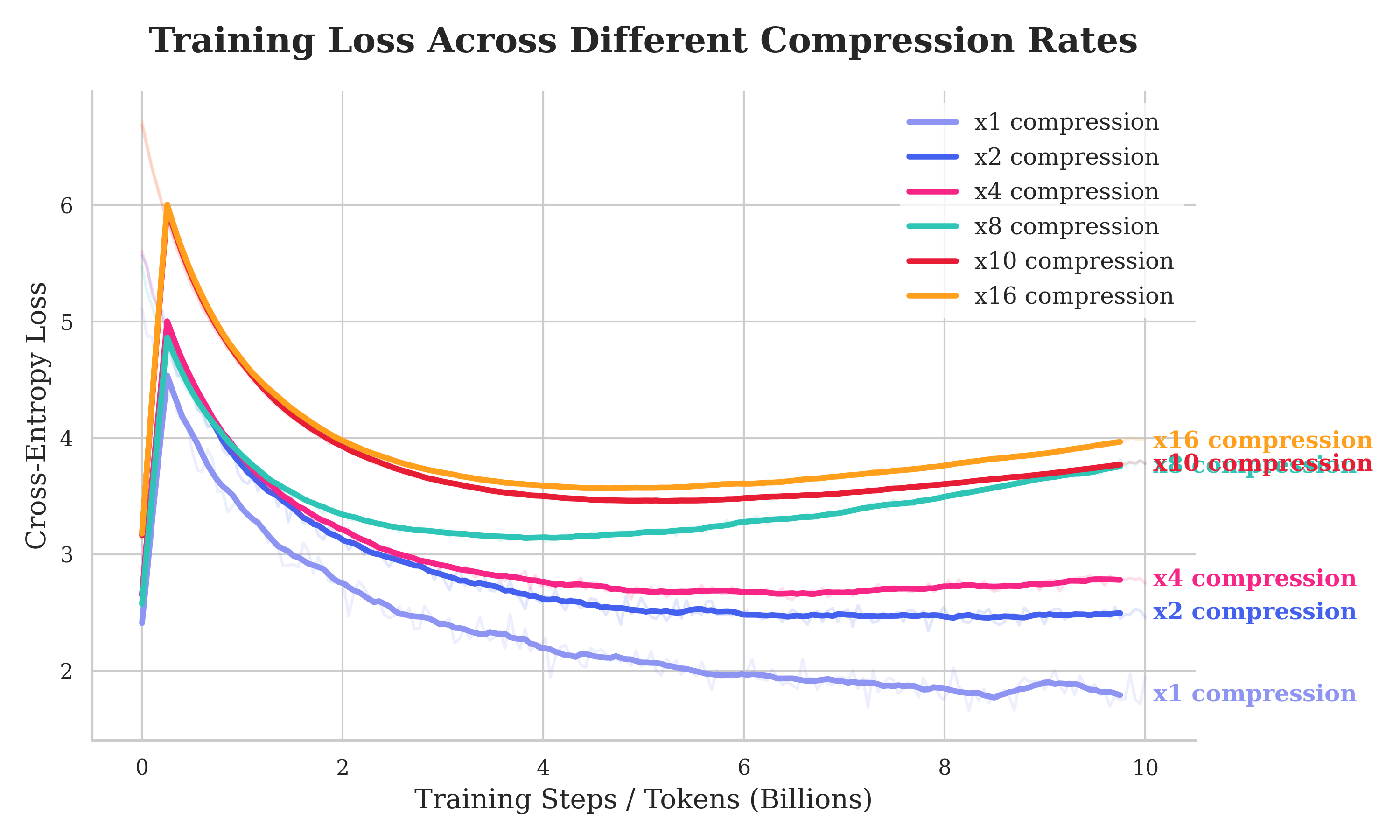}
\caption{Training loss of 0.6B model.}
\label{fig:sub1}
\end{subfigure}
\begin{subfigure}[b]{0.32\linewidth}
\includegraphics[width=\linewidth]{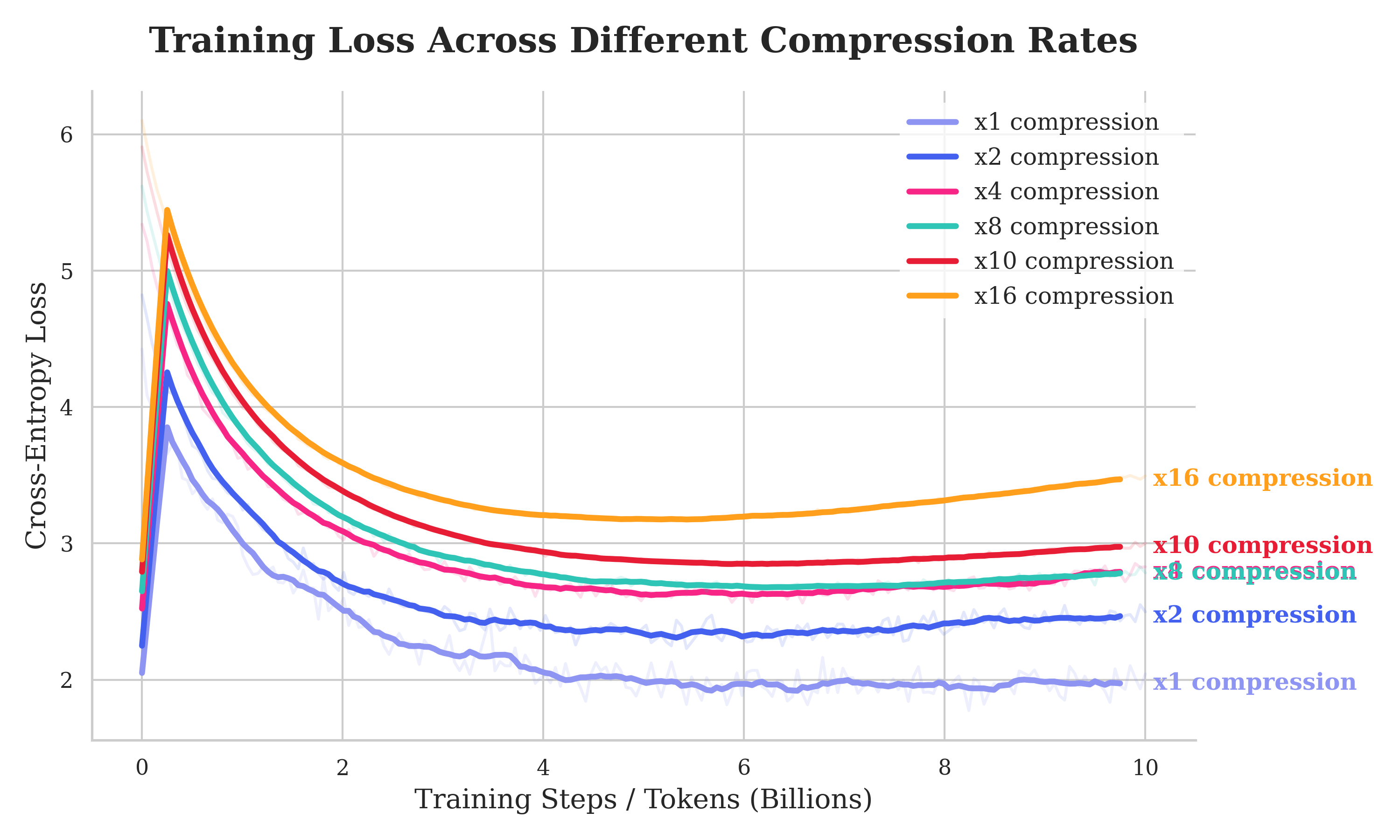}
\caption{Training loss of 4B model.}
\label{fig:sub2}
\end{subfigure}
\begin{subfigure}[b]{0.32\linewidth}
\includegraphics[width=\linewidth]{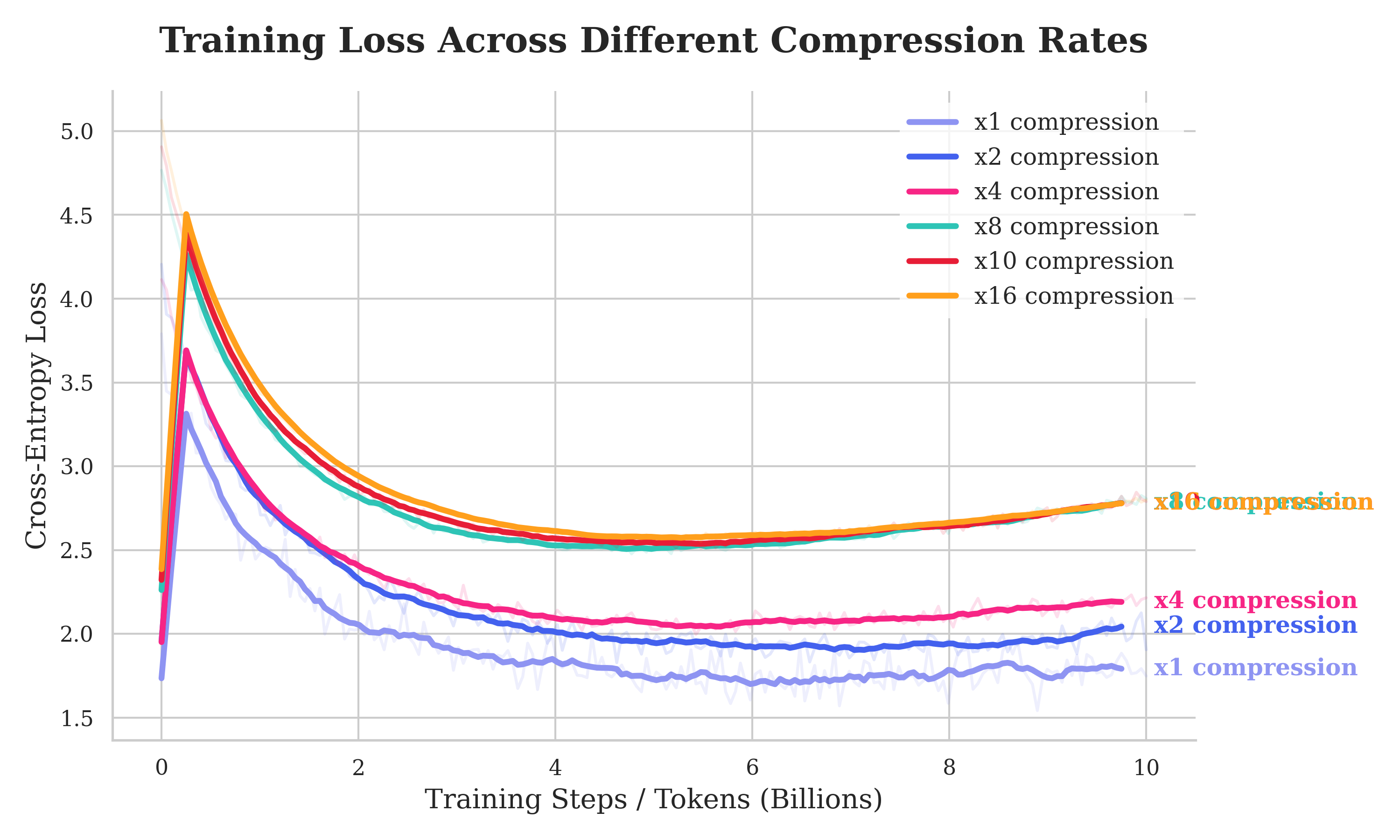}
\caption{Training loss of 8B model.}
\label{fig:sub3}
\end{subfigure}
\caption{Monitoring stage-2 pre-training loss of VIST2 models.}
\label{fig:3}
\end{figure*}

\textbf{Optical Character Recognition}.
In the second pre-training stage, we split the WuDao corpora~\cite{WuDaoCorpora} into equal-sized chunks - each consisting of $L$ continuous tokens - and render them into pure-text images. Specifically, we grid an image into 256 patches to align with the SigLIP2 configuration, which are transformed into 256 optical tokens for the VIST2 input. To find the best compression ratio $r$, we set $L\in[256,512,1024,2048,2560,4096]$ (corresponding to $r\in [1,2,4,8,10,16]$) and evaluate the OCR performance using ROUGE scores. The training runs on at most 1024$^3$ tokens, and the loss is monitored in Figure~\ref{fig:3}. 
It is noticed that a compression ratio less than 10 is not difficult for training convergence. 
However, when we evaluated the model using 1M tokens excluded from the training partition, the ROUGE-L scores in Table~\ref {tab:3} demonstrate that our comparison models suffer from unstable performance with $r>4$. As a result, we set $r=4$ in the remaining experiments to facilitate the challenging OLM training.

% \textbf{Anonymous Submission:} ICML uses double-blind review: no identifying
% author information may appear on the title page or in the paper
% itself. \cref{author info} gives further details.
\begin{table}[!t]
\centering
\scriptsize
\caption{Captioning results after stage-1 pretraining.}
\begin{tabular}{l|c|c|c|c|c}
\toprule
\textbf{Model} & \textbf{Scale} & \textbf{R-1} & \textbf{R-2} & \textbf{R-L} & \textbf{CS} \\
\midrule
Qwen3-VL & 8B & 0.309 & 0.252 & 0.307 & 0.4840 \\
InternVL-3.5 & 8B & 0.322 & 0.251 & 0.217 & 0.4863 \\
MiMo-VL & 7B & 0.327 & 0.258 & 0.318 & 0.4923 \\
Ovis2.5 & 2.2B & 0.221 & 0.168 & 0.218 & 0.3921 \\
Minicpm-4.5V & 3B & 0.308 & 0.270 & 0.302 & 0.3916 \\
GLM-4.5V & 9B & 0.335 & 0.295 & 0.328 & 0.4863 \\
\midrule
\rowcolor{gray!20} VIST2 (ours) & 0.6B & 0.462 & 0.318 & 0.451 & 0.4901 \\
% \addlinespace[1pt]
\rowcolor{gray!20} & 4B & 0.489 & 0.342 & 0.473 & 0.4881 \\
% \addlinespace[1pt]
\rowcolor{gray!20} & 8B & 0.518 & 0.369 & 0.501 & 0.4867 \\
\bottomrule
\end{tabular}
\label{tab:1}
\end{table}

\begin{table}[!t]
\centering
\scriptsize
\caption{OCR performance after stage-2 pretraining.}
\begin{tabular}{lcccccc}
\toprule
\multirow{2.5}{*}{\textbf{Model}} & \multirow{2.5}{*}{\textbf{Scale}} & \multicolumn{5}{c}{\textbf{Compression Ratio} ($r$)} \\
\cmidrule{3-7}
 & & 2 & 4 & 8 & 10 & 16 \\
\midrule
PaddleOCR-VL & 0.9B & 0.922 & 0.851 & 0.817 & 0.809 & 0.852 \\
MinerU2.5 & 1.2B & 0.967 & 0.617 & 0.619 & 0.621 & 0.568 \\
DotsOCR & 3B & 0.841 & 0.830 & 0.846 & 0.760 & 0.518 \\
Deepseek-OCR & 3B & 0.996 & 0.941 & 0.986 & 0.947 & 0.927 \\
Qwen3-VL & 8B & 0.521 & 0.368 & 0.318 & 0.370 & 0.302 \\
Qwen3-VL-SFT & 8B & 0.988 & 0.946 & 0.987 & 0.930 & 0.908 \\
\midrule
\rowcolor{gray!20} VIST2 (ours) & 0.6B & 0.972 & 0.983 & 0.938 & 0.915 & 0.890 \\
% \addlinespace[1pt]
\rowcolor{gray!20} & 4B & 0.961 & 0.942 & 0.970 & 0.928 & 0.905 \\
% \addlinespace[1pt]
\rowcolor{gray!20} & 8B & 0.981 & 0.950 & 0.975 & 0.925 & 0.912 \\
\bottomrule
\end{tabular}
\label{tab:2}
\end{table}

\begin{table}[!t]
\centering
\scriptsize
\caption{OLM performance after stage-3 pretraining.}
\begin{tabular}{clccc}
\toprule
\textbf{Scale} & \textbf{Model} & \textbf{Arxiv} & \textbf{Gutenberg} & \textbf{Wikipedia}$^{\dag}$ \\
\midrule
\textit{0.6B} & Qwen3  & 4.85 & 4.20 & 3.95 \\
% & Qwen3-VL  & \\
\rowcolor{gray!20} & VIST2 (ours) & 4.98 & 4.25 & 4.02 \\
\midrule
\textit{4B} & Qwen3 & 3.22 & 2.95 & 2.70 \\
& Qwen3-VL  & 3.48 & 3.12 & 2.82 \\
\rowcolor{gray!20} & VIST2 (ours) & 3.25 & 2.98 & 2.65 \\
\midrule
\textit{8B} & Qwen3 & 2.15 & 1.88 & 1.65 \\
& Qwen3-VL & 2.32 & 2.05 & 1.78 \\
\rowcolor{gray!20} & VIST2 (ours) & 2.22 & 1.90 & 1.62 \\
\bottomrule
\end{tabular}
\label{tab:3}
\end{table}

\subsection{Fine-tuning Performance}
\label{sec:3-2}

The modal-interleaved instruction tuning is conducted on nearly 10M samples sourced from publicly available datasets. The curation of this training set follows three key principles: 1) the samples cover both the Chinese and English languages, 2) each response prepends a chain-of-thought (CoT) within the field surrounded by <think> and </think> tags before the final answer, and 3) a response is compressed into visual tokens without differentiating CoT and answer.
After training, we test the model's multifaceted performance through extensive evaluations detailed below:
% The evaluated aspects include long-context understanding, long-text generation, and fundamental abilities.

\textbf{Long-Context Performance}. 
% We first evaluate the long-context understanding of VIST2 using the LongBench benchmark. [...] Then, we evaluate the long-text generation using the LooGLE benchmark.
We first evaluate the \textit{long-context understanding} of VIST2 using the LongBench benchmark~\cite{LongBench}. As shown in Table~\ref{tab:4}, VIST2 demonstrates a significant lead over both naive LLMs and existing compression-based methods across all sub-tasks. Specifically, VIST2-8B achieves the highest scores in QA based on single/multiple documents (45.2 and 49.8) and summarization (69.5), outperforming the strong compression baseline AdmTree. Even our smallest variant, VIST2-0.6B, delivers competitive performance in FewShot and Summ tasks, suggesting that our modal-interleaved tuning effectively preserves long-range dependency despite the aggressive visual token compression. 
Then, we evaluate the \textit{long-text generation} based on the LooGLE benchmark~\cite{LooGLE}. The results in Table~\ref{tab:5} indicate that VIST2 maintains high generation quality even with a compressed context of 8k$\times$4. In the arXiv paper summarization task, VIST2-8B achieves a GPT4 score of 88.42, which exceeds GPT4-8k (85.42) and GPT3.5-turbo (86.84). For the long dependency QA task, VIST2-8B achieves a RougeL score of 38.12 and a GPT4\_score of 56.45, showcasing its robustness in retrieving and synthesizing information across extremely long sequences. These results validate that compressing responses into visual tokens significantly enhances the ability of the backbone pure-text LLM.

\textbf{Fundamental Abilities} are evaluated based on four well-known benchmarks: GSM-8k~\cite{cobbe2021gsm8k}, MATH~\cite{MATH}, AQUA~\cite{AQUA}, and CMMLU~\cite{CMMLU}.
% Notably, optical compression is bypassed, and only textual samples are adopted in this experiment.
We compare VIST2 with naive LLMs and their visual-enhanced counterparts, and the results are reported in Table~\ref{tab:6}.
VIST2 maintains robust general-purpose intelligence across varying scales. On the mathematical reasoning benchmarks GSM-8k and MATH, VIST2-8B achieves scores of 0.87 and 0.30, respectively, slightly underperforming the baseline Qwen3-8B. This suggests that our modal-interleaved pretraining, which takes into account CoT contents in context compression, effectively maintains the model's thinking capabilities. Furthermore, VIST2-8B scores 0.75 on the Chinese comprehensive benchmark CMMLU, confirming that our bilingual curation principle preserves strong performance in non-English contexts. Overall, these findings indicate that the visual token compression strategy in VIST2 does not sacrifice fundamental LLM capacities, making it a versatile backbone for both long-document parsing and general reasoning tasks.

% \textbf{Capacity Evolution}.

\begin{table}[!t]
\centering
\caption{Comparison results on LongBench.}
\label{tab:4}
\scriptsize
\begin{tabular}{lccccc}
\toprule
\textbf{Methods} & \textbf{SglDoc} & \textbf{MtlDoc} & \textbf{Summ.} & \textbf{FewShot} & \textbf{Code} \\
% \midrule
% \multicolumn{6}{c}{\textbf{LLama-2-7B}} \\
% \midrule
% \multicolumn{6}{l}{\textit{Naive LLM}} \\
% Naive LLM & 24.7 & 22.4 & 24.6 & 63.2 & 57.7 \\
% Naive LLM-FT & 34.8 & 27.5 & 23.2 & 61.8 & 57.8 \\
% \midrule
% \multicolumn{6}{l}{\textit{Retrieval-based Methods}} \\
% BM25 & 25.1 & 23.9 & 24.4 & 56.4 & 33.1 \\
% SBERT & 17.1 & 15.8 & 23.6 & 53.2 & 36.8 \\
% OpenAI & 28.3 & 16.4 & 16.9 & 23.7 & 50.3 \\
% \midrule
% \multicolumn{6}{l}{\textit{Compression-based Methods}} \\
% AutoCompressor & 12.9 & 16.4 & 16.3 & 23.8 & 39.4 \\
% ICAEt & 19.5 & 19.2 & 19.5 & 24.8 & 27.8 \\
% LongLLMLingua & 21.5 & 18.8 & 21.7 & 49.5 & 53.2 \\
% SnapKVt & 24.2 & 22.6 & 16.3 & 60.1 & 57.7 \\
% Beacont & 34.9 & 27.5 & 25.0 & 61.4 & 57.8 \\
% AdmTree & \textbf{36.5} & \textbf{36.3} & \textbf{26.9} & \textbf{65.5} & \textbf{60.9} \\
% \midrule
% \multicolumn{6}{c}{\textbf{Qwen-2-7B}} \\
\midrule
\multicolumn{6}{l}{\textit{Naive LLM}} \\
Qwen3-4B & 25.2 & 23.8 & 23.1 & 65.5 & 55.9 \\
Qwen3-8B & 29.1 & 24.6 & 59.4 & 60.3 & 42.5 \\
\midrule
\multicolumn{6}{l}{\textit{Naive LLM-FT}} \\
Qwen3-4B & 36.4 & 39.3 & 28.9 & 68.0 & 62.7 \\
Qwen3-8B & 43.6 & 38.2 & 28.3 & 71.4 & 63.8 \\
\midrule
\multicolumn{6}{l}{\textit{Retrieval-based Methods}} \\
BM25 & 28.8 & 31.1 & 23.8 & 54.0 & 32.0 \\
SBERT & 18.4 & 22.9 & 21.6 & 50.4 & 34.6 \\
OpenAI & 30.0 & 19.1 & 23.8 & 22.9 & 51.6 \\
\midrule
\multicolumn{6}{l}{\textit{Compression-based Methods}} \\
LongLLMLinguat & 24.7 & 20.3 & 26.3 & 55.9 & 50.1 \\
SnapKV & 38.7 & 37.6 & 26.2 & 67.1 & 60.3 \\
Beacont & 40.5 & 40.3 & 26.8 & 68.4 & 66.4 \\
AdmTree (7B) & {41.6} & {45.9} & {30.2} & \textbf{69.9} & \textbf{66.8} \\
\midrule
\multicolumn{5}{l}{\textit{Ours}} \\
\rowcolor{gray!20} VIST2-0.6B & 36.8 & 39.5 & 25.2 & 54.1 & 58.4 \\
% \addlinespace[1pt]
\rowcolor{gray!20} VIST2-4B & \underline{41.5} & 44.2 & 28.1 & \underline{60.5} & 64.9 \\
% \addlinespace[1pt]
\rowcolor{gray!20} VIST2-8B & \textbf{45.2} & \textbf{49.8} & \textbf{30.4} & {65.2} & \underline{66.5} \\
\bottomrule
\end{tabular}
\end{table}

\begin{table*}[!t]
\centering
\caption{Comparison results on LooGLE.}
\label{tab:5}
\scriptsize
\begin{tabular}{lccccccccc}
\toprule
\textbf{Models} & \textbf{Context} & \textbf{Bleu1} & \textbf{Bleu4} & \textbf{Rouge1} & \textbf{Rouge4} & \textbf{RougeL} & \textbf{Meteor} & \textbf{Bert\_score} & \textbf{GPT4\_score} \\
\midrule
\multicolumn{10}{l}{\textbf{ArXiv Paper Summarization}} \\
\midrule
GPT4-32k & 32k & 24.50 & 0.73 & 27.15 & 7.10 & 24.25 & 19.03 & 84.04 & 82.84 \\
GPT4-8k & 8k & 29.02 & 2.09 & 32.08 & 11.11 & 28.85 & 22.64 & 84.92 & 85.42 \\
GPT3.5-turbo-16k & 16k & 28.70 & 1.59 & 32.04 & 10.69 & 28.89 & 22.34 & 84.82 & 86.84 \\
Llamalndex & - & 22.53 & 0.63 & 26.28 & 6.97 & 23.73 & 21.07 & 83.09 & 76.35 \\
ChatGLM2-6B & 32k & 0.04 & 0.00 & 5.97 & 0.00 & 5.82 & 6.40 & 73.25 & 13.23 \\
LongLLaMa-3B-Instruct & 256k & 0.00 & 0.00 & 0.07 & 0.00 & 0.07 & 0.33 & 26.01 & 5.57 \\
RWKV-4-14B-raven & 8k & 15.51 & 0.09 & 18.98 & 3.32 & 17.12 & 14.78 & 80.05 & 34.79 \\
LLaMA2-7B-32K-Instruct & 32k & 0.09 & 0.00 & 0.20 & 0.00 & 0.20 & 2.44 & 72.41 & 6.14 \\
\rowcolor{gray!20} VIST2-0.6B & 8k$\times$4 & 26.12 & 1.10 & 29.45 & 8.54 & 26.15 & 20.12 & 83.95 & 81.20 \\
\rowcolor{gray!20} VIST2-4B & 8k$\times$4 & 28.45 & 1.88 & 31.80 & 10.42 & 28.52 & 22.10 & 84.70 & 86.15 \\
\rowcolor{gray!20} VIST2-8B & 8k$\times$4 & 29.15 & 2.15 & 33.24 & 11.45 & 29.75 & 22.88 & 85.12 & 88.42 \\
\midrule
\multicolumn{10}{l}{\textbf{Long Dependency QA}} \\
\midrule
GPT4-32k & 32k & 8.55 & 1.40 & 25.59 & 6.36 & 24.04 & 11.13 & 80.16 & 54.09 \\
GPT4-8k & 8k & 8.94 & 1.01 & 23.45 & 6.57 & 21.69 & 10.18 & 85.36 & 42.12 \\
GPT3.5-turbo-16k & 16k & 6.92 & 1.81 & 25.02 & 6.68 & 23.63 & 10.40 & 83.79 & 45.04 \\
Llamalndex & - & 7.76 & 1.24 & 23.62 & 7.10 & 22.30 & 10.47 & 83.87 & 37.63 \\
ChatGLM2-6B & 32k & 5.55 & 0.11 & 9.41 & 1.93 & 8.69 & 4.39 & 85.78 & 11.50 \\
LongLLaMa-3B-Instruct & 256k & 5.64 & 0.49 & 17.30 & 3.76 & 16.29 & 6.53 & 84.26 & 21.64 \\
RWKV-4-14B-raven & 8k & 3.88 & 0.22 & 20.39 & 3.20 & 19.20 & 6.41 & 81.46 & 14.32 \\
LLaMA2-7B-32K-Instruct & 32k & 0.08 & 0.00 & 4.07 & 0.00 & 4.07 & 1.06 & 66.54 & 2.85 \\
Claude3-opus & 200k & 3.28 & 0.43 & 37.95 & 13.46 & 36.56 & 9.44 & 79.58 & 20.71 \\
\rowcolor{gray!20} VIST2-0.6B & 8k$\times$4 & 8.12 & 0.95 & 35.10 & 11.20 & 33.45 & 9.85 & 82.15 & 48.30 \\
\rowcolor{gray!20} VIST2-4B & 8k$\times$4 & 9.05 & 1.55 & 38.65 & 13.50 & 37.10 & 10.82 & 84.60 & 54.12 \\
\rowcolor{gray!20} VIST2-8B & 8k$\times$4 & 9.42 & 1.78 & 40.12 & 14.22 & 38.12 & 11.45 & 85.34 & 56.45 \\
\bottomrule
\end{tabular}
\end{table*}

\begin{table}[!t]
\centering
\caption{Evaluation results of general performance.}
\label{tab:6}
\scriptsize
\begin{tabular}{lcccc}
\toprule
\textbf{Methods} & \textbf{GSM-8k} & \textbf{MATH} & \textbf{AQUA} & \textbf{CMMLU} \\
\midrule
\multicolumn{5}{l}{\textit{Naive LLM}} \\
Qwen3-4B & 0.8597 & 0.2559 & 0.2840 & 0.7025 \\
Qwen3-8B & 0.8842 & 0.3120 & 0.3345 & 0.7680 \\
\midrule
\multicolumn{5}{l}{\textit{Visual-enhanced LLM}} \\
Qwen3-VL-4B & 0.7180 & 0.8583 & 0.2596 & 0.6542 \\
Qwen3-VL-8B & 0.7855 & 0.8924 & 0.2988 & 0.7135 \\
\midrule
\multicolumn{5}{l}{\textit{Ours}} \\
\rowcolor{gray!20} VIST2-4B & 0.8431 & 0.2485 & 0.2795 & 0.6950 \\
\rowcolor{gray!20} VIST2-8B & 0.8715 & 0.3015 & 0.3210 & 0.7512 \\
\bottomrule
\end{tabular}
\vspace{-10pt}
\end{table}

% \begin{table}[!t]
% \centering
% \caption{Evaluation results of general performance.}
% \label{tab:7}
% \scriptsize
% % \setlength{\tabcolsep}{5pt}
% \begin{tabular}{lcccc}
% \toprule
% \textbf{Methods} & \textbf{R-Bench} & & & \\
% \midrule
% % \multicolumn{5}{c}{\textbf{LLama-2-7B}} \\
% % \midrule
% \multicolumn{5}{l}{\textit{Naive VLM}} \\
% Qwen3-VL-4B & \\
% Qwen3-VL-8B & \\
% \midrule
% \multicolumn{5}{l}{\textit{Naive VLM-FT}} \\
% Qwen3-VL-4B & \\
% Qwen3-VL-8B & \\
% \midrule
% \multicolumn{5}{l}{\textit{Ours}} \\
% VIST2-4B & \\
% VIST2-8B & \\
% \bottomrule
% \end{tabular}
% \end{table}

\textbf{Optical Language Modeling}.
OLM requires our VIST2 to continue the content of spaced chunks conditioned solely on their optical representations. To achieve this, we gather long documents (exceeding 8k tokens) from three corpora: Arxiv~\cite{ArXiv}, Gutenberg\footnote{\url{https://www.gutenberg.org/}}, and WuDao~\cite{WuDaoCorpora}, to create a training set. 
After training, we evaluate the resulting model by providing the first 1k tokens of a document and using the perplexity calculated by GPT-4 to quantify model performance in writing the remaining content (at most 4k output tokens). Results in Table~\ref{tab:3} indicate that: 
1) The existing foundational vision-language model is inadequate for the continuation of long texts.
2) After the training with OLM, the long-writing performance of VIST2 surpasses the VLM encounters by 0.2$\sim$0.5 on average, which is close to the baseline pure language models.

\subsection{Efficiency Analysis}
\label{sec:3-3}
\begin{figure}[!t]
\centering
\includegraphics[width=0.9\linewidth]{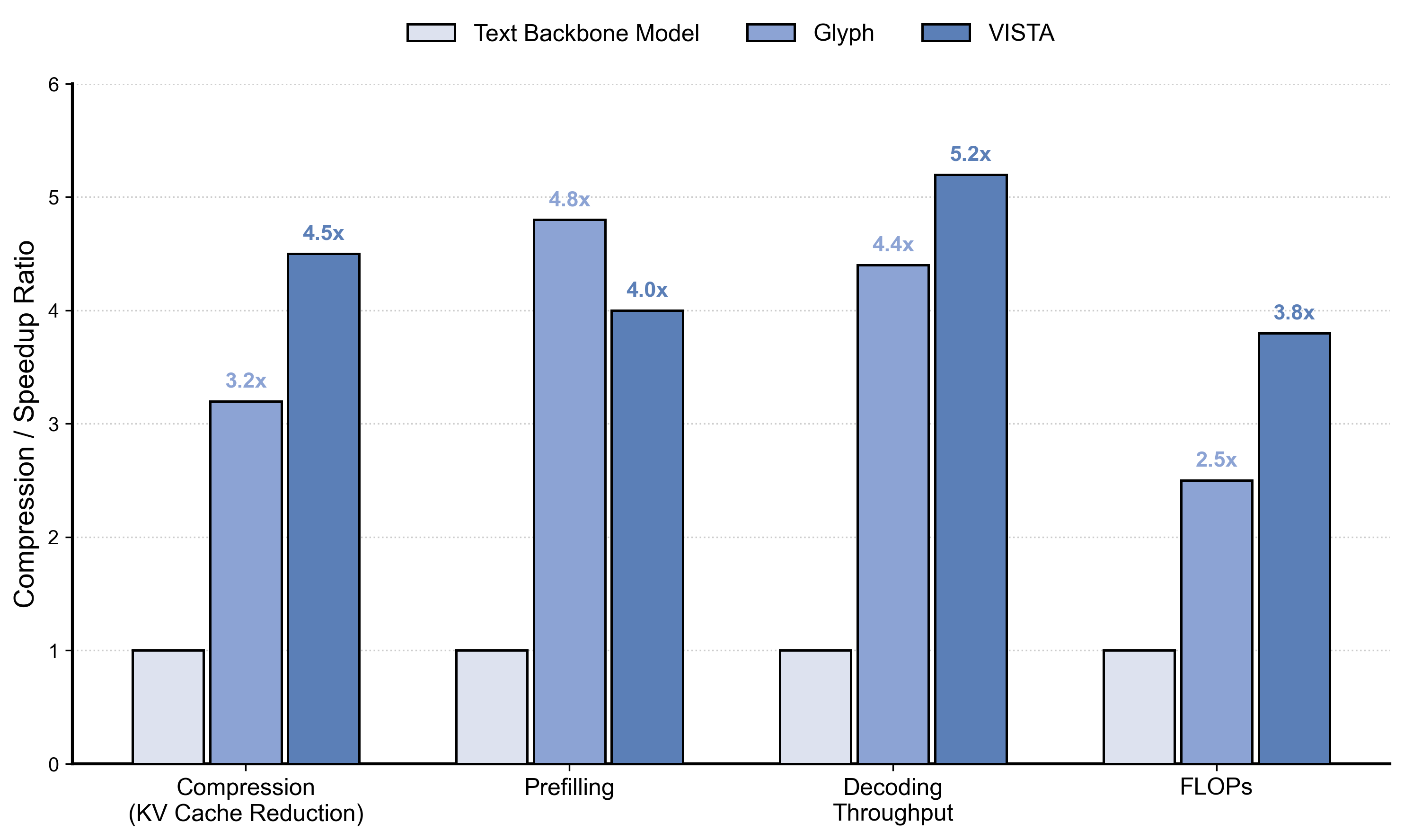}
\caption{Comparison of efficiency.}
\vspace{-10pt}
\label{fig:5}
\end{figure}

In Figure~\ref{fig:5}, we compare VIST2 with another optical context compression approach, Glyph, in terms of savings in computing and memory cost, as well as improvements in responding speed and throughput. 
Note that VIST2 and Glyph share a similar compression ratio ($\sim4\times$), and the two models represent PCC and GCC, respectively. GCC presents a significant advantage in KV-Cache saving and FLOPs reduction. The two models have an equivalent throughput, and VIST2 achieves a slightly lower prefilling compression because it is implemented with a fixed compression ratio, without an enhanced visual encoder or an adaptive compression mechanism, which is left for future work.

\section{Conclusion}
\label{sec:5}
In this work, we conduct a pioneering investigation of global context compression (GCC). Additionally, we propose VIST2, a novel Transformer architecture that achieves GCC with interleaved vision-text transformation. The key technique of VIST2 is a staged training recipe that connects the advanced modal-interleaved instruction tuning and the basic visual-language tasks with optical language modeling. As a result, VIST2 presents higher efficiency in saving computing and memory costs than previous PCC approaches, while maintaining advances in long text understanding and generation. 

\section{Future Works}
\label{sec:6}

While VIST2 demonstrates a substantial leap in efficient optical-based context compression, several promising avenues for future research remain:
\begin{itemize}[leftmargin=*]  % , nosep
\item \textbf{Specialized Visual Encoders for Texts}. In this work, VIST2 utilizes a general-purpose visual encoder that has not been specifically optimized for the high-density textual information found in documents. Future iterations could explore the integration of document-centric visual models (e.g., specialized CLIP-style variants trained on academic or structured text) to achieve even higher semantic compression ratios without compromising granular details.
\item \textbf{Content-Aware Adaptive Compression}. Our current framework employs a static compression law based on fixed-size chunking. However, document regions vary significantly in information density (e.g., blank margins vs. complex tables). Implementing an adaptive compression mechanism - one that dynamically allocates visual tokens based on the local informativeness or structural complexity of each chunk - could further optimize the trade-off between computational efficiency and reconstruction fidelity.
\item \textbf{Modal-interleaved Reinforcement Learning}. This study focuses on the pre-training and supervised fine-tuning of VIST2. A natural next step is to investigate alignment techniques based on reinforcement learning, specifically tailored for modal-interleaved contexts. Such alignment could better harmonize the reasoning process with the verifiable rewards. 
\end{itemize}

\bibliography{custom}

\newpage
\appendix
\label{sec:appendix}

\begin{table*}[!t]
\centering
\scriptsize
\caption{
Details of the multi-stage training of VIST2.
\dag: datasets for long-text understanding.
\ddag: datasets for long-text generation.
}
\begin{tabular}{cp{3.5cm}cccc}
\toprule
\multirow{2.5}{*}{\textbf{Task}} & \multirow{2.5}{*}{\textbf{Dataset}} & \multicolumn{3}{c}{\textbf{Frozen}} & \multirow{2.5}{*}{\#\textbf{Data Volume}} \\
\cmidrule{3-5}
& & \textbf{VE} & \textbf{MC} & \textbf{LLM} & \\
\midrule
\multicolumn{6}{l}{\textit{Pre-training}} \\
\midrule
Caption & CoCo, SA1B & \ding{51} & \ding{55} & \ding{51} & 690M Images \\
MT-OCR & WuDao & \ding{55} & \ding{55} & \ding{51} & 1B Tokens \\
OLM & Arxiv, Gutenberg, WuDao & \ding{51} & \ding{51} & \ding{55} & 1B Tokens \\
\midrule
\multicolumn{6}{l}{\textit{Fine-tuning}} \\
\midrule
Instruct-Follow & Arxiv$^{\dag}$, Gutenberg$^{\dag}$, NarrativeXL$^{\dag}$, WuDao$^{\dag}$, Deepseek-R1-Distill$^{\ddag}$ & \ding{51} & \ding{51} & \ding{55} & 10M Instructions \\
\bottomrule
\end{tabular}
\label{tab:data}
\end{table*}

\begin{table}[!ht]
\centering
\scriptsize
\caption{
Hyperparameter settings. 
PT: pre-training.
SFT: supervised instruction-tuning.
$^{\dag}$: batch size achieved through gradient accumulation.
}
\label{tab:hp}
\begin{tabular}{lcccc}
\toprule
\textbf{Stage} & \textbf{PT-1} & \textbf{PT-2} & \textbf{PT-3} & \textbf{SFT} \\
\midrule
% \multicolumn{5}{c}{\textbf{Training}} \\
% \midrule
Optimizer & \multicolumn{4}{c}{AdamW} \\
Weight Decay & \multicolumn{4}{c}{1e-4} \\
Warmup Ratio & \multicolumn{4}{c}{0.01} \\
Learning Rate Schedule & \multicolumn{4}{c}{Cosine} \\
\# of Epochs & \multicolumn{4}{c}{1} \\
Learning Rate & 5e-4 & 5e-4 & 5e-4 & 1e-5 \\
% Minimal Learning Rate & \multicolumn{3}{c}{5e-5} \\
Batch Size (VIST2-0.6B) & 96 & 64 & 64$^{\dag}$ & 64$^{\dag}$ \\
Batch Size (VIST2-4B) & 64 & 32 & 16$^{\dag}$ & 16$^{\dag}$ \\
Batch Size (VIST2-8B) & 48 & 24 & 8$^{\dag}$ & 8$^{\dag}$ \\
% \# of Training Samples & 154,582 & 32,747 & 5,976 \\
% \# of Validation Samples & 2,048 & 3,461 & 1,496 \\
\# of Maximum Length & 1,024 & 4,096 & 8,192 & 8,192 \\
\bottomrule
\end{tabular}
\end{table}

\section{Experimental Settings}
\subsection{Image Captioning}
To evaluate the model performance of image captioning, we compare VIST2 with six naive VLMs: 

\noindent \textbf{Qwen3-VL}~\cite{bai2025qwen3vltechnicalreport}: Qwen3-VL family utilizes an enhanced interleaved-MROPE for spatial-temporal modeling and supports native interleaved contexts of up to 256K tokens for long-context comprehension of documents and videos.

\noindent \textbf{InternVL-3.5}~\cite{InternVL3.5}: Featuring a Cascade Reinforcement Learning framework and a Visual Resolution Router, this open-source series achieves state-of-the-art results by balancing advanced reasoning capabilities with high inference efficiency.

\noindent \textbf{MiMo-VL}~\cite{coreteam2025mimovltechnicalreport}: MiMo-VL is a multimodal model designed to bridge the gap between open-source and commercial models through advanced reinforcement learning techniques focused on general multimodal and agentic tasks.

\noindent \textbf{Ovis2.5}~\cite{lu2025ovis25technicalreport}: Ovis2.5 is a high-performance vision-language model that employs a structural alignment strategy to better process high-resolution images and complex visual reasoning tasks.

\noindent \textbf{Minicpm-4.5V}~\cite{yu2025minicpmv45cookingefficient}: MiniCPM-V 4.5 is a versatile, end-side multimodal model that provides strong OCR and multimodal understanding capabilities while maintaining a compact parameter size for efficient deployment.

\noindent \textbf{GLM-4.5V}~\cite{GLM-4.5V}: GLM-4.5V is a large-scale multimodal model that excels in multidisciplinary reasoning and high-resolution document understanding, rivaling leading commercial models in its perception and generation quality.

\subsection{Optical Character Recognition}
To evaluate the model performance of optical character recognition, we compare VIST2 with four VLMs tailored for OCR, apart from the naive Qwen3-VL and the Qwen3-VL fine-tuned on our dataset: 

\noindent \textbf{PaddleOCR-VL}~\cite{cui2025paddleocrvlboostingmultilingualdocument}: This ultra-compact 0.9B parameter model integrates a NaViT-style dynamic resolution visual encoder with the ERNIE-4.5-0.3B language model to achieve state-of-the-art efficiency in multilingual document parsing across 109 languages.

\noindent \textbf{MinerU2.5}~\cite{MinerU}: MinerU2.5 employs a decoupled two-stage framework that separates global layout analysis from local content recognition, enabling high-resolution parsing of complex elements like formulas and tables with minimal computational overhead.

\noindent \textbf{DotsOCR}~\cite{dots.ocr}: This unified 1.7B-parameter vision-language model is designed to jointly learn layout detection, content recognition, and relational understanding within a single end-to-end pass, providing robust performance on the XDocParse benchmark.

\noindent \textbf{Deepseek-OCR}~\cite{DeepSeek}: Deepseek-OCR introduces the concept of "context optical compression," utilizing a multi-stage DeepEncoder and a Mixture-of-Experts decoder to compress 2D document pages into a compact set of vision tokens for efficient high-accuracy transcription.

\subsection{Fine-tuning Performance}
For the assessment of long-context understanding and long-text generation, we compared VIST2 against models that have publicly available evaluation results.
Results in Table~\ref{tab:4} are reported in~\cite{li2025admtree}, and results in Table~\ref{tab:5} are reported in~\cite{LooGLE}, respectively.

\section{Implementation Details}
\noindent \textbf{Datasets}. 
Table~\ref{tab:data} presents the datasets utilized in the VIST2 training process, categorized by stages. 

At the fine-tuning stage, note that Deepseek-R1-Distill\footnote{\url{https://huggingface.co/datasets/a-m-team/AM-DeepSeek-R1-0528-Distilled}} is already an instruction-following dataset; we filter samples with responses longer than 4,096 tokens to inflate our training set. 
Additionally, we employ NarrativeXL~\cite{narrativexl}, a long document comprehension dataset with the letter of the ground-truth option as the target label. We augment it by asking Qwen3\footnote{\url{https://huggingface.co/Qwen/Qwen3-235B-A22B-Thinking-2507}} to answer the questions after thinking. Only the samples with a format of thought-analysis-option and longer than 2,048 tokens are incorporated into our training data.
For the Arxiv, Gutenberg, and WuDao corpora, we extract documents longer than 512 tokens and prompt Qwen3 to generate questions varying in difficulty levels (1 to 5). Subsequently, we instruct Qwen3 to provide answers using a chain-of-thought approach for each document-question pair. Response texts exceeding 2,048 tokens are incorporated into our dataset. During training, we randomly sample 10 million entries from this resulting dataset for modal-interleaved instruction tuning.

\noindent \textbf{Training Configurations}.
Our codes are implemented with Pytorch 2.6.0 and the Huggingface Transformers repository. 
Table~\ref{tab:hp} further reports the training configurations.

\section{Application Examples}
Figure~\ref{fig:e1} illustrates the usage of VIST2, where the short query remains uncompressed while the lengthy response is effectively compressed into the visual context during the generation process. In addition, Figure~\ref {fig:e2} demonstrates that the long query is also compressed during context pre-filling. The original 3,784 tokens are reduced to 1,024 visual tokens, resulting in significant savings in the KV-Cache.

\section{Ablation of OLM Training}
\begin{figure}[!ht]
\centering
\includegraphics[width=0.95\linewidth]{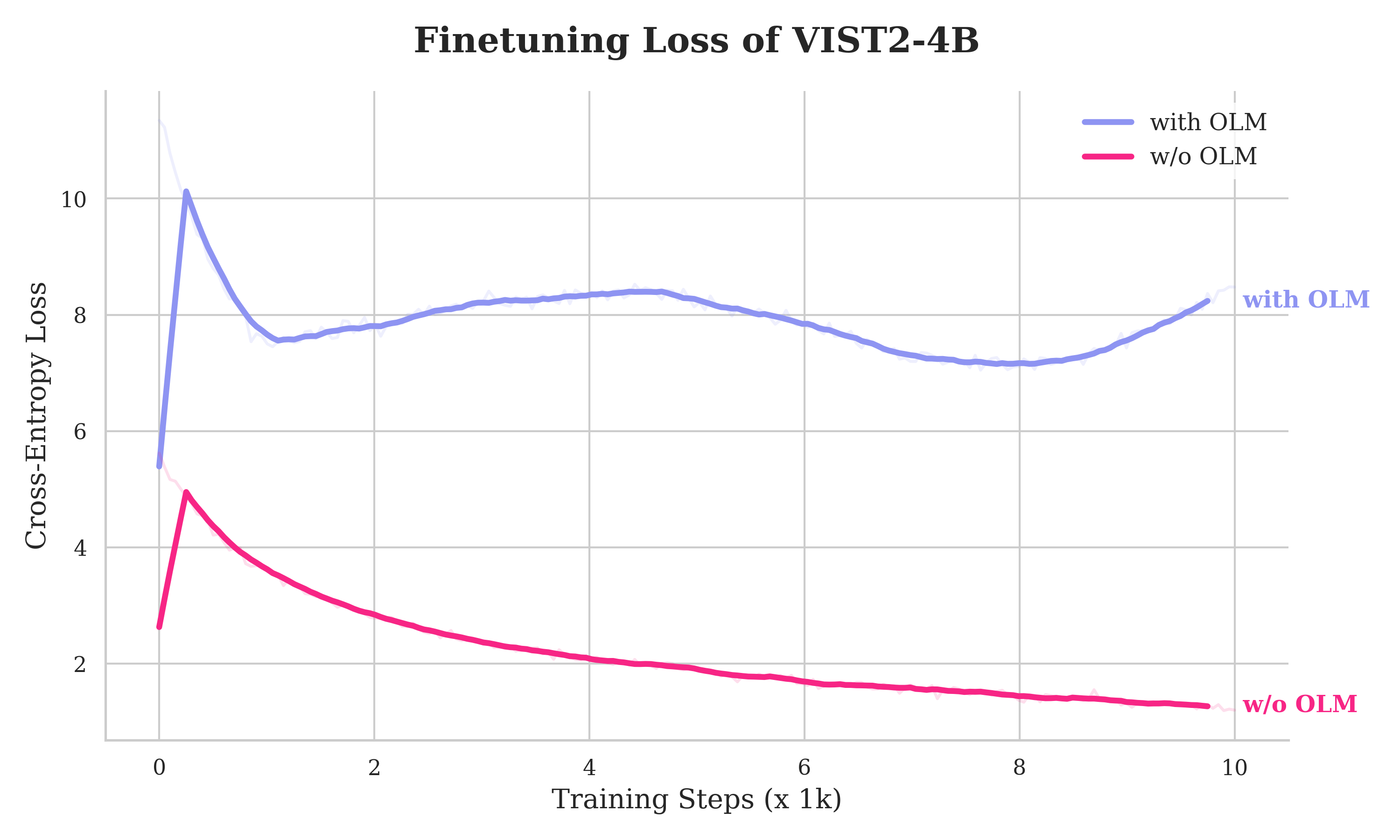}
\caption{Training loss of image captioning.}
\label{fig:loss}
\end{figure}

\begin{figure*}[t]
\centering
\small
% 定义表格列格式：c(文字居中)
\begin{tabular}{cc} % Adjusting to a standard c for center alignment
% 第一行
\rotatebox{90}{\textbf{Unfinished Response}} & 
\adjincludegraphics[width=0.9\linewidth,valign=c]{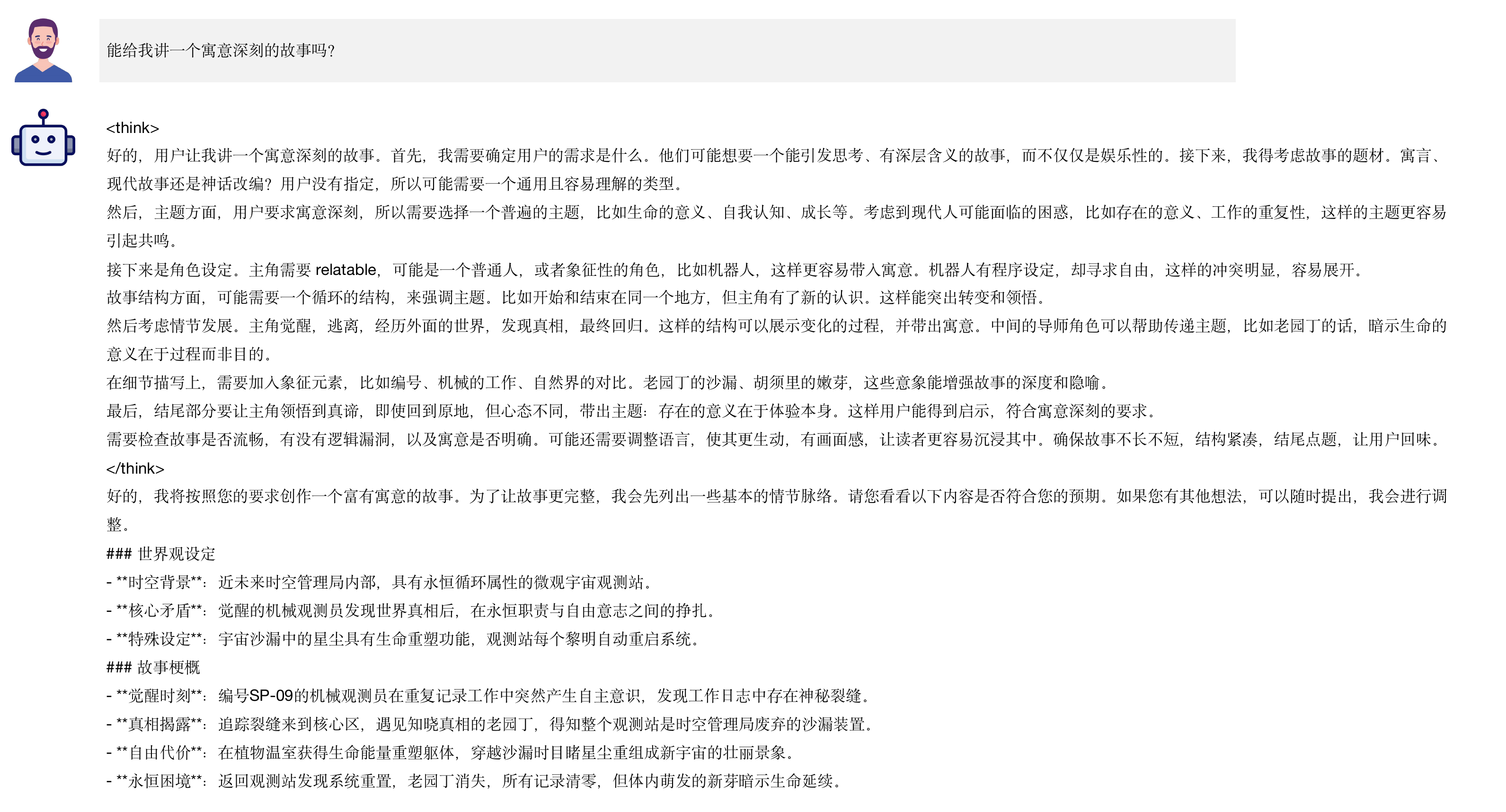} \\
\noalign{\smallskip} % 行间距微调
% 第二行
\rotatebox{90}{\textbf{Continue Response}} & 
\adjincludegraphics[width=0.9\linewidth,valign=c]{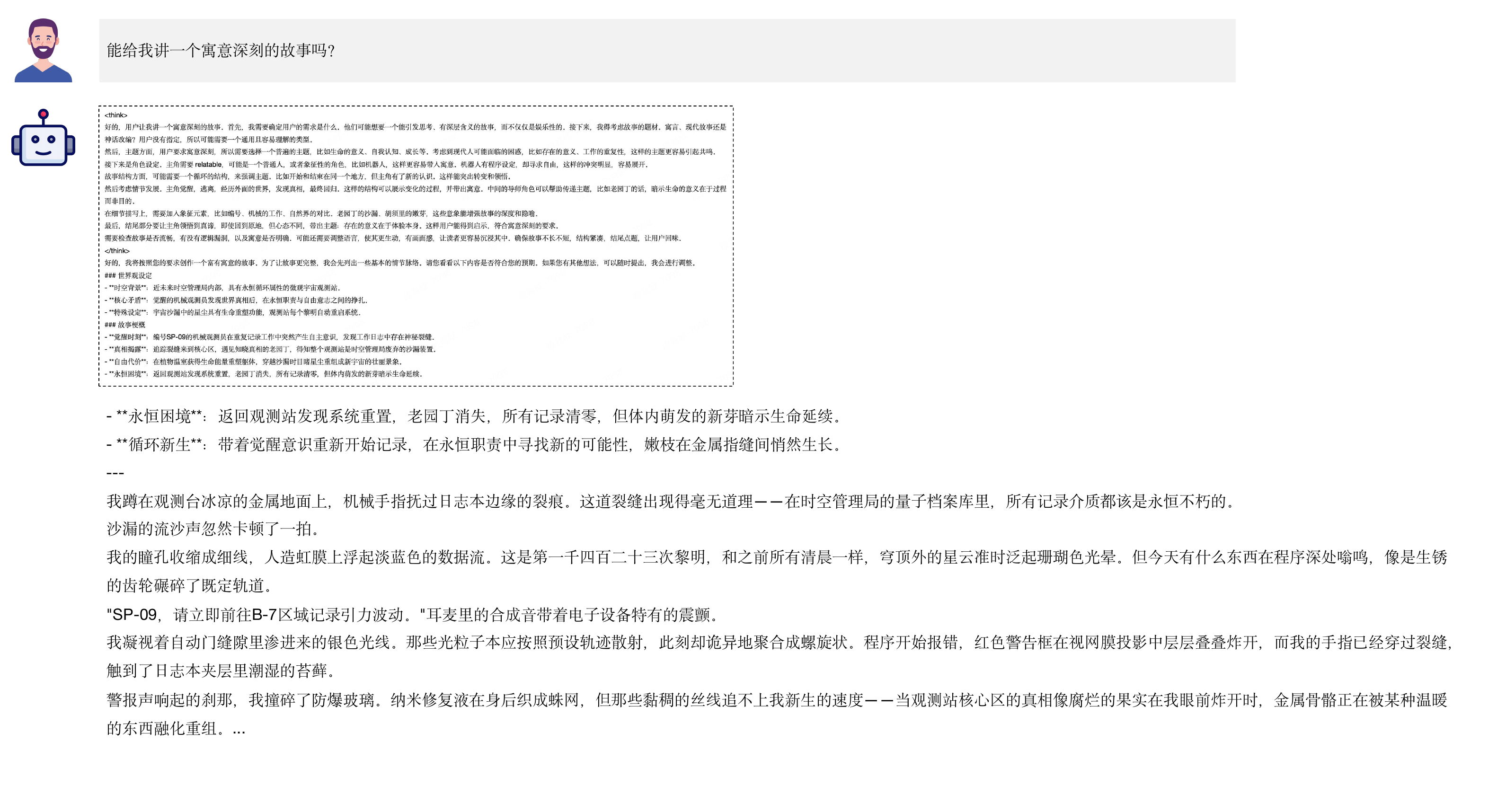} \\
\noalign{\smallskip}
% 第三行
\rotatebox{90}{\textbf{Continue Response}} & 
\adjincludegraphics[width=0.9\linewidth,valign=c]{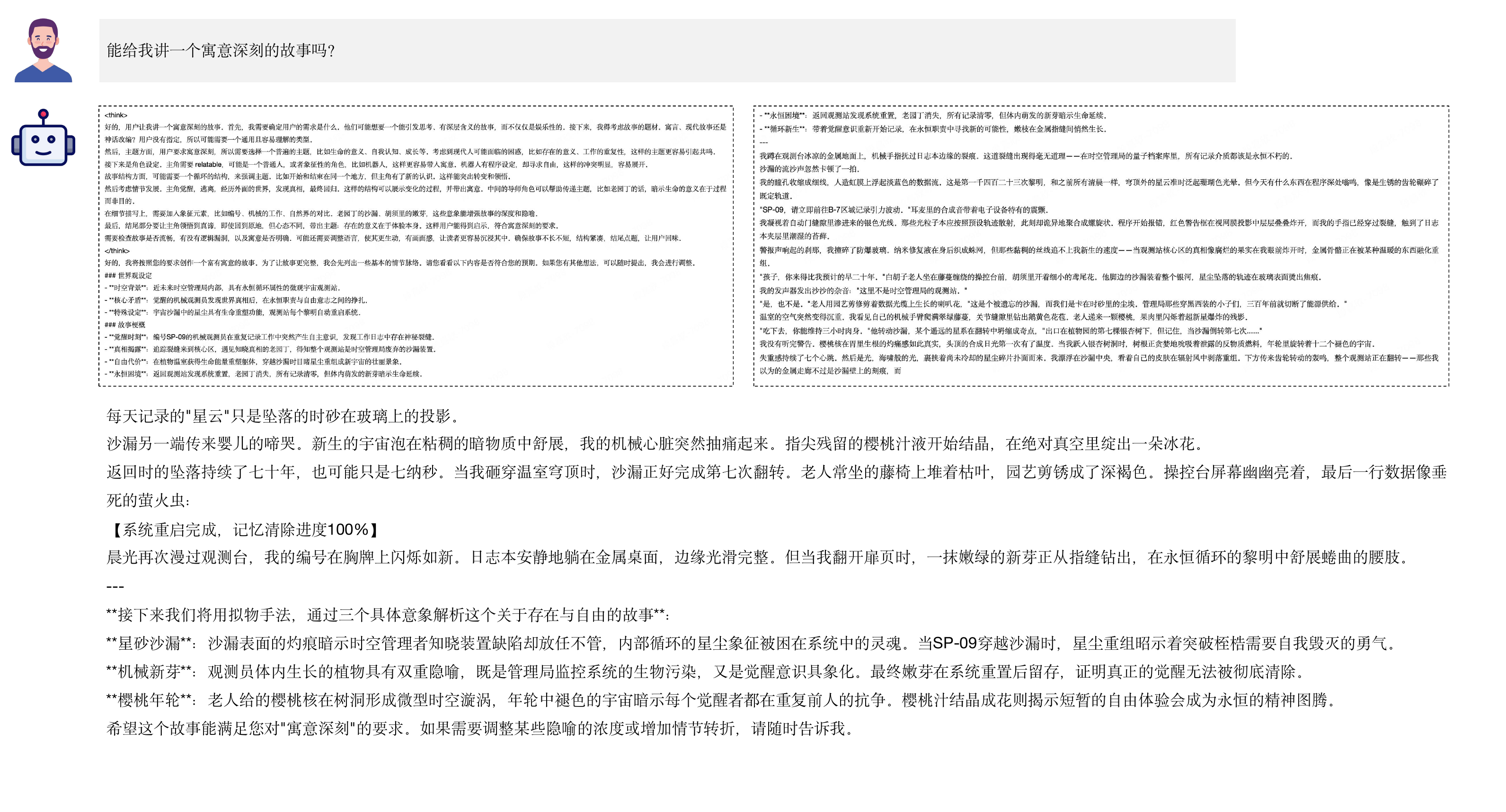} \\
\end{tabular}
\caption{A testing example with short query and long response.}
\label{fig:e1}
\end{figure*}

\begin{figure*}[t]
\centering
\small
\includegraphics[width=1.0\linewidth]{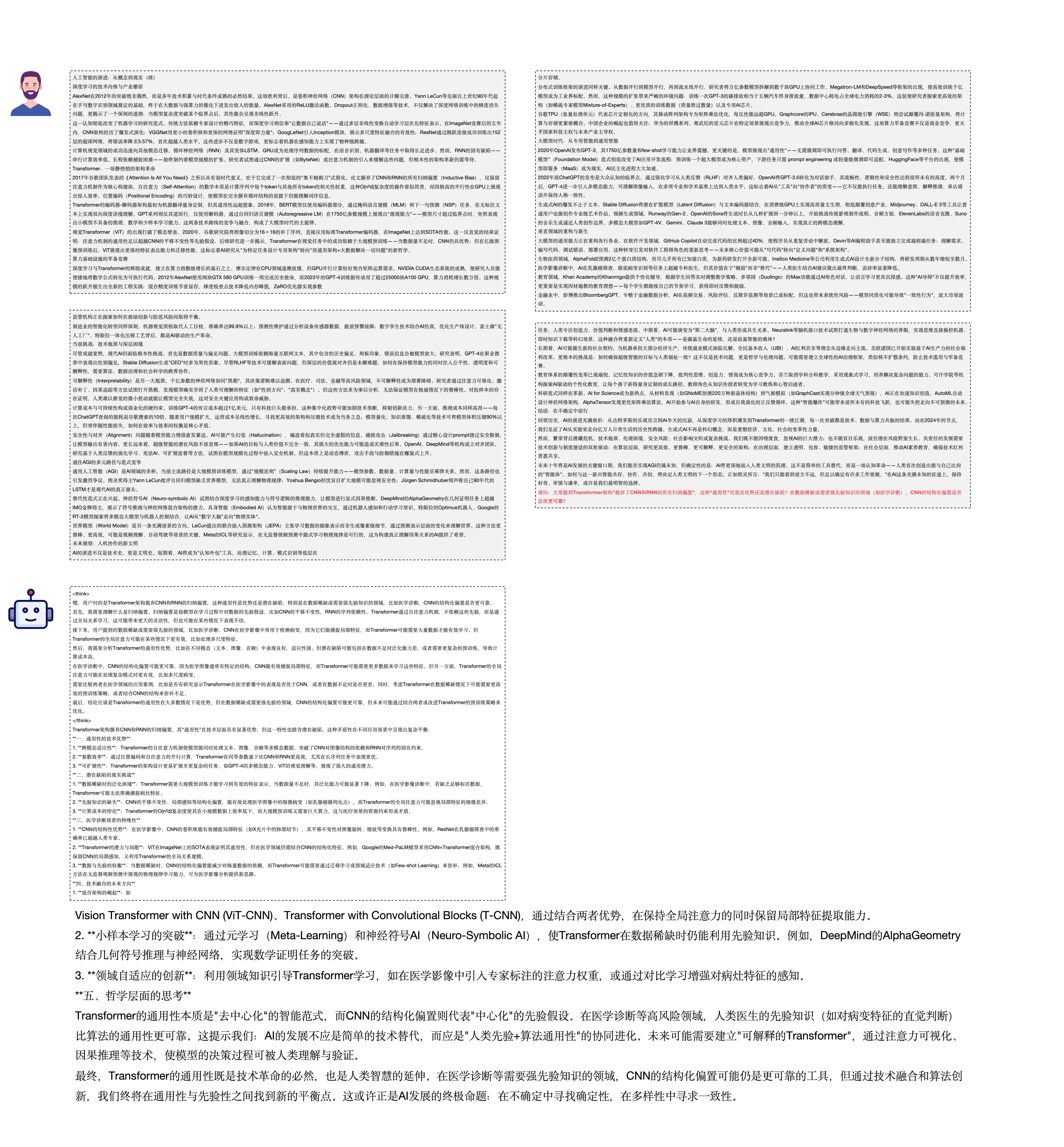}
\caption{
A testing example with a long query and a long response.
The user query is rendered with a \colorbox{gray!20}{grey background} to differentiate from the assistant response, and the user's question is \textcolor{red}{red} to differentiate from the document.
The reading progression follows a left-to-right, top-to-bottom order.
}
\label{fig:e2}
\end{figure*}

In Figure~\ref{fig:loss}, we examine the necessity of stage-3 pre-training with OLM by monitoring the fine-tuning loss of the VIST2-4B variants. The results indicate that intermediate modal-interleaved instruction tuning using OLM significantly improves training stability, resulting in a smoother convergence of loss.

% \begin{figure}
% \centering
% \includegraphics[width=0.95\linewidth]{images/improved_comparison.png}
% \caption{Training loss of image captioning.}
% \label{fig:6}
% \end{figure}

% \begin{figure*}[!ht]
% \centering
% \includegraphics[width=0.95\linewidth]{images/training_stages_comparison.png}
% \caption{Basic capacity evolution of VIST2 alongside training stages.}
% \label{fig:4}
% \end{figure*}

\end{document}